# Machine Learning and the Internet of Things Enable Steam Flood Optimization for Improved Oil Production

Mi Yan
ExxonMobil Global Service Company
mi.yan1@exxonmobil.com

Jonathan C. MacDonald
Imperial Oil Resources Ltd.
jonathan.c.macdonald@esso.ca

Chris T. Reaume
ExxonMobil Global Service Company
chris.t.reaume@esso.ca

Wesley Cobb
ExxonMobil Global Service Company
wesley.cobb@exxonmobil.com

Tamas Toth
ExxonMobil Global Service Company
tamas.toth@exxonmobil.com

Sarah S. Karthigan
ExxonMobil Global Service Company
sarah.karthigan@exxonmobil.com

## ABSTRACT

Recently developed machine learning techniques, in association with the Internet of Things (IoT) allow for the implementation of a method of increasing oil production from heavy-oil wells. Steam flood injection, a widely used enhanced oil recovery technique, uses thermal and gravitational potential to mobilize and dilute heavy oil *in situ* to increase oil production. In contrast to traditional steam flood simulations based on principles of classic physics, we introduce here an approach using cutting-edge machine learning techniques that have the potential to provide a better way to describe the performance of steam flood. We propose a workflow to address a category of time-series data that can be analyzed with supervised machine learning algorithms and IoT. We demonstrate the effectiveness of the technique for forecasting oil production in steam flood scenarios. Moreover, we build an optimization system that recommends an optimal steam allocation plan, and show that it leads to a 3% improvement in oil production. We develop a minimum viable product on a cloud platform that can implement real-time data collection, transfer, and storage, as well as the training and implementation of a cloud-based machine learning model. This workflow also offers an applicable solution to other problems with similar time-series data structures, like predictive maintenance.

**ACM Reference format:**

## 1 INTRODUCTION

### 1.1 Background

Machine learning [1, 2] and the Internet of Things (IoT) [3-8] have proven successful across various industries. They have recently gained more attention in the oil and gas industry. Machine learning offers an alternative solution to a number of traditional problems in the oil and gas industry [9-15]. Bergen *et al.* [9] summarized applications of machine learning in geoscience. Xu *et al.* [15] reviewed recent progress in petrophysics using machine learning. The development of IoT helps achieve real-time data acquisition via embedded sensors, as well as model building and its deployment at IoT edge devices or a cloud platform. It has a wide range of applications in the oil and gas industry, particularly in the upstream industry, the oil exploration and production sector. Khan *et al.* [16] proposed a reliable and efficient IoT-based architecture for the oilfield environment, and Aalsalem *et al.* [17] presented a review of recent advances and open challenges in wireless sensor networks in the oil and gas industry. By combining the capabilities of machine learning and IoT, we propose an effective method to forecast oil production in steam flood scenarios as well as a steam-allocation optimization system to predict a potential 3% increase in oil production.

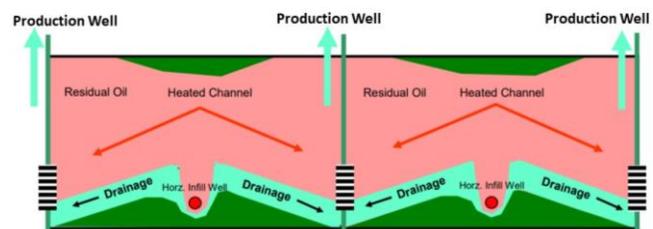

**Figure 1: Steam flood process.**

Ways of increasing oil production to meet growing global energy demands is a popular subject of research. Enhanced oil recovery (EOR) [18] is a widely used technique, which helps increase oil production in the post-natural-extraction process. Usually the natural pumping stage results in as much as 70% residual crude oil owing to low well pressure. To efficiently recover oil, three primary EOR methods—thermal injection [19], gas injection, and chemical injection—are carried out depending on field conditions. Steam flood injection [20-22] is a major thermal injection technique in which steam is injected into infill wells to mobilize and dilute heavy oil using thermal and gravitational potential, so that production wells can easily extract oil from reservoirs (Fig. 1).

### 1.2 Related work

Oil production forecast in steam flood fields has been studied for decades. Traditional analytical models [21, 22] were built using principles of physics and reservoir conditions to describe the performance of the steam flood and predict oil production with the given steam-allocation plans (Fig. 2). However, there were considerable discrepancies between actual oil production and the predictions. Few machine-learning-based studies of steam flood injection have been reported in the literature. Hama *et al.* [23] employed hierarchical clustering, an unsupervised machine learning algorithm, to create a new steam flood screening criterion to help choose the method of EOR to use for different reservoir conditions. Nevertheless, it does not cover oil production forecasting.

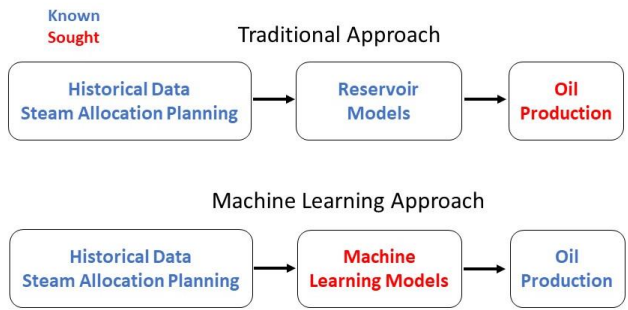

**Figure 2: Comparison of traditional and machine learning approaches to oil production forecasting.**

### 1.3 How can machine learning and IoT help?

In the traditional method, an output, i.e., oil production, is predicted by known models and input data. Machine learning methods offer a different way to solve this problem. In the machine learning approach, the input and output in a training dataset are used to train candidate models (Fig. 2). The optimal model is determined according to predefined criteria and evaluated on a test dataset that has never been seen before. In other words, machine learning methods are data driven whereas traditional approaches are physics-principles driven. In contrast to traditional techniques, geological parameters are not necessary for building machine learning models. Moreover, some decision-tree-based machine learning algorithms, e.g., Random Forests [24] and XGBoost [25], can help recognize the relative importance of each factor in a complex non-linear system. With these powerful capabilities, machine learning can help a steam-flood surveillance team quantitatively explore opportunities to improve oil production.

Machine learning and IoT are complementary. Machine learning can process a massive amount of data collected by IoT edge devices to reveal hidden patterns that are difficult to be identified otherwise, whereas IoT allows for real-time data acquisition and storage, as well as the machine learning model training and implementation on a cloud platform or IoT edge devices. This work benefits from the development of both machine learning and IoT.

## 2 DATA AND METHODS

### 2.1 Data collection and introduction

The raw data are collected by edge sensors in the field from five data sources. The sources have different schematic design, primary keys, and sampling frequency. Daily extract, transform and load jobs are used to collect and cross reference the data, perform error correction, and consolidate them to one location. They are then transferred to a cloud storage to be ready for the data engineering process.

Table 1 shows the structure of a daily dataset at the well level in a pad composed of a group of adjacent dependent wells. Our goal is to build one model for one pad to forecast the daily oil production of each production well. There are 16 features in the dataset. *Well Name* has two categories: *infill wells* and *production wells*. *Sensor Data* contains the real-time measurements of temperature and pressure in the field. *Steam Volume* is the daily steam volume injected into each infill well, which is only valid for *infill wells*, whereas *Well Status*, *Sensor Data* and *Oil Volumes* are only meaningful for *production wells*. *Oil Volume* is the daily oil production of each production well, which is the output variable. As expected, missing data are inevitable in the empirical dataset. Before feeding data into machine learning algorithms, the data engineering stage is necessary.

| *Date* | *Well Name* | *Well Status* | *Sensor Data* | *Steam Volume* | *Oil Volume* |
|---|---|---|---|---|---|
| **4/17/2019** | Prod Well 1 | Pump | 100 | NA | 23 |
| **4/18/2019** | Prod Well 1 | Shut-In | 200 | NA | 31 |
| … | … | … | … | … | … |
| **4/17/2019** | Infill Well 1 | NA | NA | 6 | NA |
| **4/18/2019** | Infill Well 1 | NA | NA | 9 | NA |
| … | … | … | … | … | … |

**Table 1: Structure of a daily dataset at the well level.**

## 2.2 Workflow

As illustrated in Figure 3, there are five sections in the workflow: data collection and transfer, data engineering, model building, data visualization, and the optimization system. We perform missing data imputation using forward copy and backward copy for different features. No individual record is dropped, as a complete history of wells is needed for the data engineering stage. As introduced before, there are two categories of wells, and so the dataset is separated into two subsets accordingly. Using the infill-well subset, we create a new data structure, where each row corresponds to daily records, and columns/features are the daily steam volumes injected into each infill well. To the production-well subset, two groups of features are added. *gas_day_rate* considers the effective working time of pumps, and one-hot encodings of the production wells are derived from categorical features, e.g., *Well Name* and *Well Status*. We merge the reorganized infill-well subset with the new production-well subset aligned by *date* to build a new dataset.

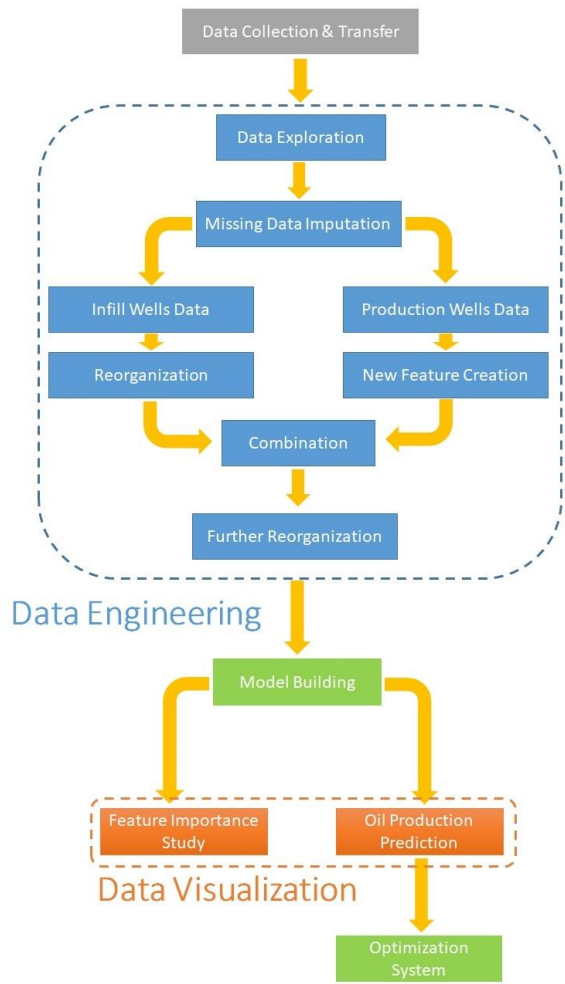

**Figure 3: Schematic of workflow of steam flood optimization.**

There are two questions in this case: (1) What is the future oil production with a given steam-allocation plan? (2) How does historical data influence future production? For a $t$-day-prior daily production prediction with $k$-day historical data as input, we create two groups of features to account for the two questions, respectively:

*prior_m_day_infill_well_x_steam*,
where $m = 1, 2, …, t - 1, t$, with infill well names $x$,
&
*prior_n_day_sensor_y_value*,
where $n = t + 1, …, t + k$, with sensor names $y$.

The difference in the ranges of $m$ and $n$ occurs because $m$ focuses on the impact of future steam plans whereas $n$ indicates those of historical records. Note that $k$ is a hyperparameter depending on the data. Until this point, the input dataset has been ready for building daily production forecast models at the well level, with typical dimensions of 100,000 rows by 1000 columns/features.

All models are trained with the GPU-supported XGBoost algorithm [25]. The optimal model is finalized by the hyperparameter grid search using $k$-fold time-series cross-validation.

## 3 RESULTS

### 3.1 Importance Study

We show here the results of an optimal 30-day-prior prediction model. Figure 4 lists the eight most important features given by the model. A surveillance team can gain from this a qualitative impression of how important each feature is, which can help them with planning and decision making.

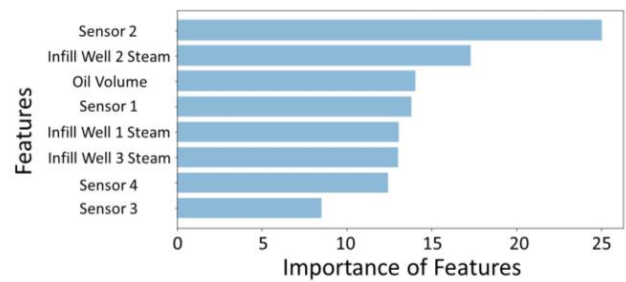

**Figure 4: Importance study results (top 8) of the optimal model.**

### 3.2 Oil Production Forecast

The dataset is split into training (80%) and test (20%) datasets. Figure 5 compares the normalized monthly real oil production with the predicted production. A ±10% relative error band with respect to real production is displayed for reference. The real daily production and prediction are both at the well level, accumulated by *month* over all production wells in one pad for ease of visualization. The optimal model is selected from the five-fold time-series cross-validation in terms of the metric of *root mean square error* (RMSE). Predictions on the test dataset are all within the ±10% range of relative error of the real productions, which is a significant improvement compared with previous works [21, 22]. Table 2 summarizes the performance of the optimal model and baseline model on the training and test datasets. The baseline model predicts the future daily oil production by copying the latest real daily production, i.e. the 30-day-prior daily production here. The XGBoost model significantly outperforms the baseline model in terms of both RMSE and $R^2$.

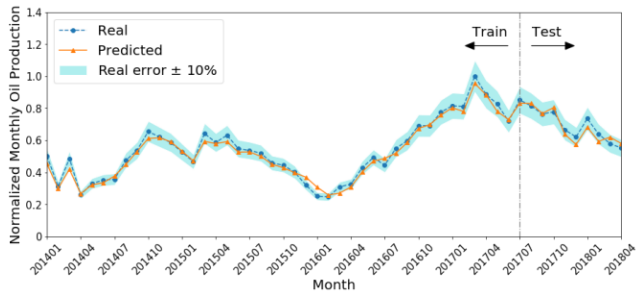

**Figure 5: Comparison of real production with predictions of the optimal 30-day-prior XGBoost model.**

| Metric | Train | Test |
|---|---|---|
| **RMSE** | 2.375 | 3.334 |
| **Baseline RMSE** | 4.295 | 4.361 |
| $R^2$ | 0.805 | 0.631 |
| **Baseline** $R^2$ | 0.361 | 0.368 |

**Table 2: Comparison of performance of the optimal model and the baseline model on the training and test datasets.**

### 3.3 Optimization System

With a model capable of accurately predicting oil production in different scenarios, optimizing steam flood allocation is straightforward. Using a pad with three infill wells as an example, the entire volume of the steam volume was injected into Infill Well 2, and the total real monthly oil production was 4202 m$^3$ (Table 3). Given this input, the model predicts oil production of 4242 m$^3$, 0.9% higher than the real value. To recommend an optimal steam allocation plan to maximize production, a brutal-force search of all possible scenarios is a simple solution. Considering a fixed total steam volume and three infill wells, there are two independent parameters, e.g., steam volumes injected into Infill Well 1 and Infill Well 2. Figure 6 plots the oil production as a function of relative steam volumes into Infill Well 1 and Infill Well 2 in percentage. The bottom-right dashed circle corresponds to the actual scenario, while the left circle indicates the optimal scenario of 27% steam injected into Infill Well 1, 4% to Infill Well 2, and 69% into Infill

Well 3, where the model predicts a maximum oil production of 4340 m$^3$ (Table 3), a 3.3% improvement compared with the real production.

In future work, given that the time needed for brutal-force search exponentially increases with the number of infill wells, other optimization algorithms, e.g., gradient descent search, can be used for pads with a large number of infill wells. Moreover, this optimization system is built with the objective function of maximizing oil production. If other objective functions are defined, such as ones to minimize the steam-oil ratio to save on fuel cost, a different optimization strategy can be employed.

|  | Real | Optimal |
|---|---|---|
| Infill Well 1 | 0% | 27% |
| Infill Well 2 | 100% | 4% |
| Infill Well 3 | 0% | 69% |
| Real Oil Production | 4202 | NA |
| Predicted Oil Production | 4242 | 4340 |

**Table 3: Comparison between real and optimal scenarios.**

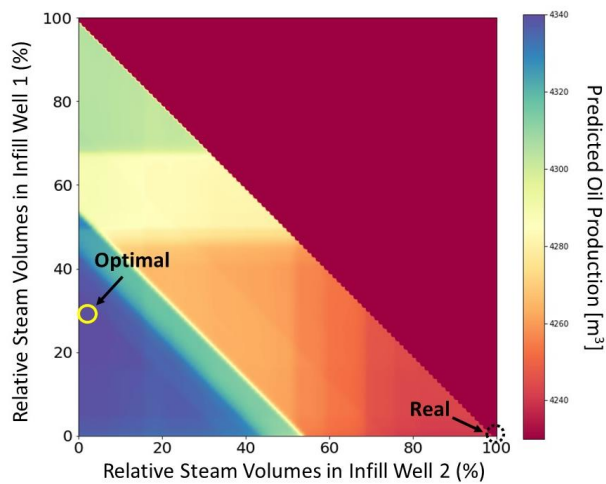

**Figure 6: Variation of oil production with relative steam volumes injected into Infill Well 1 and Infill Well 2.**

## 4 CONCLUSION

In this paper, we have designed a workflow containing the data engineering process to address a category of time-series data and a machine learning algorithm, XGBoost. This model can predict oil production in specific steam flood scenarios with unprecedented accuracy, especially compared to traditional methods. Furthermore, we build an optimization system that can recommend the optimal steam allocation plan with a 3% uplift in oil production. Benefiting from the development of cloud platforms for IoT, we develop a cloud-based minimum viable product for steam flood optimization that can achieve the real-time data collection, transfer, and storage, as well as the training and implementation of the machine learning model on cloud platforms. This work offers new opportunities for studying the steam flood in the oil and gas industry.

It will be interesting to explore applications of this workflow in other datasets with similar time-series data structures. For example, in a predictive maintenance case [26, 27], given historical records and the status of future work, this workflow can help build a machine learning model to predict the quantified level of abrasion of a machine part.

## ACKNOWLEDGMENTS

The authors would like to thank Charles A. Crawford, Richard J. Smith, Ryan C. Bouma, Dustin J. Bespalko, and Brian P. Vu for helpful discussions.